\documentclass[conference]{IEEEtran}
\IEEEoverridecommandlockouts
\usepackage{cite}
\usepackage{amsmath,amssymb}
\usepackage{amsmath,amssymb,amsfonts}
\usepackage{algorithmic}
\usepackage{graphicx}
\usepackage{textcomp}
\usepackage{xcolor}
\usepackage{booktabs}
\usepackage{multirow}
\usepackage{hyperref}
\usepackage{url}
\usepackage{algorithm}
\usepackage{balance}
\usepackage{diagbox}
\newcommand{\best}[1]{\textbf{#1}}
\newcommand{\second}[1]{\underline{#1}}

\def\BibTeX{{\rm B\kern-.05em{\sc i\kern-.025em b}\kern-.08em
    T\kern-.1667em\lower.7ex\hbox{E}\kern-.125emX}}
\begin{document}

\title{TopoEvo: A Topology-Aware Self-Evolving Multi-Agent Framework for Root Cause Analysis in Microservices  \\
\thanks{}
}
\author{\IEEEauthorblockN{1\textsuperscript{st} Junle Wang}
\IEEEauthorblockA{\textit{School of Artificial Intelligence,} \\
\textit{Beihang University}\\
Beijing, China \\
wjl\_admire@buaa.edu.cn}
\and
\IEEEauthorblockN{2\textsuperscript{nd} Xingchuang Liao}
\IEEEauthorblockA{\textit{School of Artificial Intelligence,} \\
\textit{Beihang University}\\
Beijing, China \\
liaoxingchuang@buaa.edu.cn}
\and
\IEEEauthorblockN{3\textsuperscript{rd} Wenjun Wu}
\IEEEauthorblockA{\textit{School of Artificial Intelligence,} \\
\textit{Beihang University}\\
Beijing, China \\
wwj09315@buaa.edu.cn}
\and
}
\maketitle

\begin{abstract}
Root cause analysis (RCA) in microservices is challenging due to (i) noisy and heterogeneous multimodal observability (metrics, logs, traces), (ii) cascading failure propagation that amplifies downstream symptoms, and (iii) non-stationary topology drift induced by autoscaling and rolling updates. Recent LLM-based RCA agents can generate tool-grounded explanations, yet they often remain topology-agnostic and suffer from \emph{symptom-amplification bias}, misattributing the root cause to salient downstream victims.
We propose \textbf{TopoEvo}, a topology-aware self-evolving multi-agent framework that couples graph representation learning with structured, topology-constrained reasoning.
TopoEvo first introduces \emph{Metric-orthogonal Multimodal Alignment} (MOMA), which decomposes metric embeddings into complementary subspaces and contrastively aligns logs and traces to reduce modality redundancy and sparsity, yielding stable node representations for graph encoding.
It then applies \emph{Vector Quantization} (VQ) to discretize topology-enhanced states into auditable \emph{symptom tokens} with a symptom lexicon, enabling reliable retrieval and token-level evidence grounding.
On top of these discrete topology cues, TopoEvo performs a multi-agent \emph{Hypothesis--Evidence--Test} (HET) workflow to explicitly verify propagation-consistent explanations and separate initiating anomalies from amplified downstream symptoms.
Finally, a \emph{Self-Evolving Mechanism} refreshes hierarchical incident memory and performs conservative test-time adaptation with high-confidence pseudo-labels to maintain robustness under drift.

We evaluate TopoEvo on both a public AIOps benchmark and a real-world production incident dataset. Compared to the state-of-the-art baselines, TopoEvo achieves absolute improvements of up to 3.44\% in root cause localization accuracy, while remarkably boosting fault-type classification performance by 4.39\% to 16.81\% across diverse datasets.
\end{abstract}


\section{Introduction}
Microservice architectures have become a dominant paradigm for building large-scale cloud applications due to their flexibility, support for independent deployment, and rapid iteration.\cite{soldani2021survey,wang2024survey,pham2025rcaeval}
However, this architectural shift also makes modern systems increasingly complex to operate: failures rarely stay local, and subtle issues can quickly propagate along service dependencies, producing amplified and misleading symptoms downstream\cite{yao2024coe,pan2023dycause}. 
As a result, \emph{root cause analysis} (RCA)---identifying the initiating faulty entity and its fault type---has become a critical capability for maintaining service reliability and meeting strict QoS/SLA requirements.

Despite substantial progress, existing RCA solutions still face three practical limitations in real microservice deployments.
\textbf{First}, microservices generate \emph{multimodal observability} (metrics, logs, traces), yet many approaches underutilize this richness or rely on simplistic fusion\cite{wu2021mmrca,li2024mrca,lou2023unimodal}.
In particular, heterogeneous modalities differ in frequency, sparsity, noise patterns, and missingness, making \emph{effective cross-modal alignment} non-trivial; naive concatenation often yields unstable representations and spurious correlations.
\textbf{Second}, with the rise of LLMs, multi-agent diagnosis has emerged as a promising direction for RCA, enabling tool-grounded inspection and human-readable explanations. 
However, most agent-based pipelines remain \emph{topology-agnostic}: they struggle to internalize microservice dependency constraints and therefore tend to over-trust the most salient downstream symptoms (symptom amplification), leading to inefficient investigation and frequent misattribution\cite{wang2024rcagent,pei2025flowofaction,zhang2024mabc,tang2025microrcaagent}. \textbf{Third}, microservice environments are inherently non-stationary: autoscaling, rolling updates, and evolving call graphs introduce distribution shifts and out-of-distribution (OOD) behaviors. 
Static RCA models and fixed reasoning heuristics often degrade sharply under such drift, lacking a reliable mechanism to refresh knowledge and adapt\cite{zheng2025ocean,phan2024microcercl}.

To address these challenges, we propose \textbf{TopoEvo}, a topology-aware, reasoning-enhanced, and self-evolving framework for joint \emph{microservice root cause localization} and \emph{fault type classification}. 
TopoEvo systematically connects representation learning with topology-constrained reasoning:
(1) a \emph{metric-anchored orthogonal multimodal alignment} module constructs stable shared subspaces for heterogeneous signals, mitigating modality sparsity and redundancy;
(2) a \emph{topology-aware enhancement} module discretizes topology-aware states via vector quantization (VQ) and builds a \emph{symptom vocabulary}, turning opaque vectors into compact, retrievable, and auditable evidence units for reasoning;
(3) a \emph{reasoning-enhanced multi-agent} workflow follows an explicit hypothesis--evidence--test loop under topology constraints, reducing symptom-amplification errors and improving diagnostic efficiency; and
(4) a \emph{self-evolving mechanism} continuously refreshes incident knowledge and cautiously adapts the encoder under drift, improving robustness to OOD configurations.

Our main contributions are summarized as follows:
\begin{itemize}
    \item \textbf{Metric-anchored orthogonal multimodal alignment.}
    We propose a metric-centric alignment scheme that decomposes metric embeddings into \textbf{orthogonal} subspaces and aligns logs and traces to complementary components via contrastive learning, mitigating modality sparsity and reducing redundant correlations in multimodal observability.

    \item \textbf{Topology-aware symptom tokenization for agent reasoning.}
    We introduce a VQ-based discretization of topology-aware graph states and construct a \emph{symptom vocabulary} that maps discrete codes to compact, auditable symptom tokens, enabling topology-constrained reasoning and reducing symptom-amplification misattribution in LLM-based RCA.

    \item \textbf{Reasoning-enhanced multi-agent RCA.}
We propose a hypothesis--evidence--test (HET) multi-agent workflow that explicitly models fault propagation paths and performs tool-grounded evidence verification, yielding more reliable and explainable RCA under noisy multimodal telemetry.

    \item \textbf{Self-evolving adaptation under drift.}
We introduce a self-evolving mechanism that refreshes incident knowledge and continuously adapts the graph encoder and alignment objectives under distribution shift, improving robustness to dynamic microservice configurations and out-of-distribution (OOD) incidents.
\end{itemize}

\section{Preliminaries and Motivation}

\subsection{Observability of Microservice}
Microservice observability refers to the capability of inferring internal system states and diagnosing runtime issues from external signals emitted by distributed services.
In practice, observability data are commonly summarized as three pillars: \emph{metrics}, \emph{logs}, and \emph{traces}.

\textbf{Metrics} are time-stamped numerical measurements (e.g., QPS, latency percentiles, error rates, CPU/memory usage) that provide compact and continuous views of service health and resource consumption.

\textbf{Logs} are discrete event records produced by services and infrastructure components, typically containing semi-structured messages, levels, and contextual fields; they capture rich semantic clues about failures but are often noisy and heterogeneous.

\textbf{Traces} record end-to-end request executions across services, usually organized as a trace graph of \emph{spans} with parent--child and causal relationships; they expose cross-service propagation paths and fine-grained latency breakdowns.
Given an incident time window, we denote the multimodal observability of node $v$ as
$\mathcal{O}_v=\{\mathbf{x}_v^{metric}, \mathbf{x}_v^{log}, \mathbf{x}_v^{trace}\}$,
where $\mathbf{x}_v^{metric}$ is a metric time-series segment, $\mathbf{x}_v^{(log}$ is a log snippet/set, and $\mathbf{x}_v^{trace}$ is the trace-derived feature set\cite{li2021tracerca,jha2022cloud}.

\subsection{Vector Quantization \& Codebook.}
\label{sec:VQ}
Vector Quantization (VQ) maps continuous embeddings into a finite set of discrete prototypes, enabling compact and symbolic representations\cite{chillarege1992odc}.
Given a feature vector $\mathbf{z}\in\mathbb{R}^{d}$, VQ assigns it to the nearest entry in a learnable codebook
$\mathcal{C}=\{\mathbf{c}_k\}_{k=1}^{K}$ by
\begin{equation}
k^*=\arg\min_{k\in[K]}\ \|\mathbf{z}-\mathbf{c}_k\|_2,\qquad 
\mathrm{VQ}(\mathbf{z})=\mathbf{c}_{k^*}.
\end{equation}
The codebook can be viewed as a dictionary of representative patterns that summarize frequently occurring structures in the embedding space.
During training, VQ modules are typically optimized with a reconstruction-style objective (e.g., VQ-VAE), which pulls codebook entries toward the assigned vectors while encouraging encoder outputs to commit to selected prototypes.
A common formulation is
\begin{equation}
\mathcal{L}_{\mathrm{VQ}}=
\underbrace{\|\mathrm{sg}[\mathbf{z}]-\mathbf{c}_{k^*}\|_2^2}_{\text{codebook loss}}
+\beta\underbrace{\|\mathbf{z}-\mathrm{sg}[\mathbf{c}_{k^*}]\|_2^2}_{\text{commitment loss}},
\end{equation}
where $\mathrm{sg}[\cdot]$ denotes the stop-gradient operator and $\beta$ controls the strength of the commitment term.
By discretizing dense vectors into code indices, VQ yields interpretable ``tokens'' and can reduce sensitivity to noise, which is useful when exposing learned latent patterns to downstream reasoning modules.

\subsection{Motivation}
\label{sec:motivation}

Microservice incidents are rarely isolated. 
A fault triggered at an upstream component often propagates along the dependency graph, gradually amplifying observable symptoms (e.g., retries, queueing, and timeouts) at downstream services. 
This propagation nature makes RCA fundamentally a \emph{topology-conditioned} inference problem: the most anomalous node is not necessarily the initiating cause, and correct diagnosis requires reasoning over multi-level entities (node--service--pod) and their interactions.

\subsubsection{Motivation 1}Topology-unaware LLM-based RCA suffers from symptom-amplification bias.
\label{sec:motivation_topology}
\begin{figure}[t]
  \centering
  \includegraphics[width=\columnwidth]{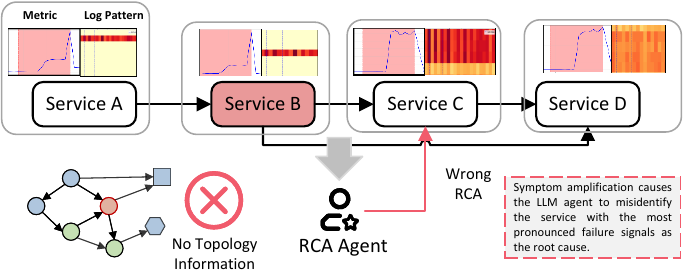}
  \caption{Illustration of symptom-amplification bias.}
  \label{fig:moti}
\end{figure}
Recent LLM-based RCA systems have shown promising capabilities in tool use and explanation generation, yet most of them remain largely \emph{topology-agnostic}. 
They typically ingest multimodal observations and produce conclusions by pattern matching or narrative reasoning, without enforcing structural constraints from the microservice dependency graph.
This omission leads to a systematic failure mode: \emph{symptom-amplification bias}(shown in Fig.~\ref{fig:moti}). 
When latency and error signals accumulate along a call chain, downstream services may exhibit the strongest symptoms, so an unconstrained reasoning process tends to select the most salient downstream node as the root cause, even though it is merely a victim of upstream propagation.

This challenge is exacerbated by the \emph{multi-level} nature of microservice systems. 
A root cause may originate from a pod-level resource contention, a service-level misconfiguration, or a node-level failure, while the observable “peak anomaly” may appear at a different level. 
Therefore, simply providing an LLM with raw logs/traces/metrics is insufficient: the model needs an explicit mechanism to \emph{perceive} and \emph{operate on} hierarchical topology evidence.
This motivates TopoEvo to couple a GAT-based localizer with topology-aware vector quantization (VQ) and a symptom lexicon, transforming dense topology-aware states into discrete, retrievable symptom tokens.
These tokens serve as compact evidence units that allow the reasoning layer to stay anchored to propagation structure, explicitly separating \emph{initiating anomalies} from \emph{amplified downstream symptoms}.

\subsubsection{Motivation 2}Dynamic topology requires reasoning-enhanced adaptation under OOD drift.
\label{sec:motivation_ood}
Microservice environments are highly non-stationary.
Autoscaling continuously adds/removes pods, rolling updates change service versions, and configuration changes alter call patterns and dependency edges.
Such dynamics induce distribution shift in both graph topology and observability, causing static RCA models to degrade---even when the fault semantics remain similar.
In practice, the same fault type may manifest differently after a topology change, and previously reliable patterns can become out-of-distribution (OOD).

This motivates an RCA design that is robust to topology drift.
On one hand, LLM-based reasoning provides a natural advantage: it can test hypotheses, reconcile incomplete evidence, and generalize beyond exact pattern matches.
On the other hand, reasoning alone is not enough for sustained performance: the system must \emph{retain} validated diagnostic knowledge and \emph{adapt} to new topologies quickly.
Therefore, TopoEvo introduces a reasoning-enhanced multi-agent workflow (hypothesis--evidence--test) to explicitly verify causal explanations under topology constraints, and a self-evolving mechanism that (1) refreshes hierarchical incident memory and (2) performs conservative test-time adaptation using high-confidence pseudo-labels.
Together, these mechanisms enable TopoEvo to leverage strong OOD reasoning while continuously aligning its encoder and knowledge base with evolving service dependencies.
\section{Methodology}
\label{sec:method}
\begin{figure*}[t]
  \centering
  \includegraphics[width=\textwidth]{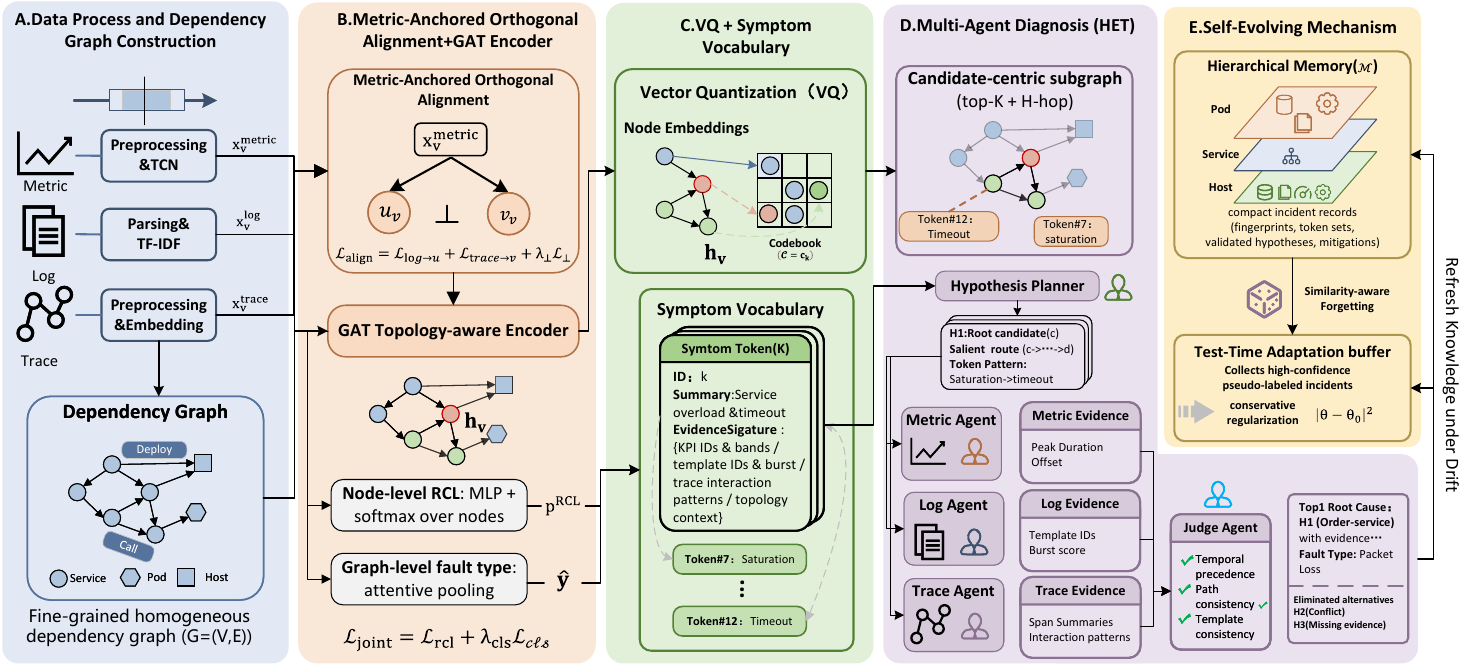}
  \caption{The overview of TopoEvo.}
  \label{fig:twocol}
\end{figure*}

As shown in Fig.\ref{fig:twocol}, TopoEvo consists of 5 components that support joint microservice root cause localization and fault type classification: 
(1) data process and dependency graph construction
(2) metric-orthogonal multimodal alignment on a fine-grained dependency graph, 
(3) topology-aware enhancement via vector quantization and symptom vocabulary, 
(4) reasoning-enhanced multi-agent diagnosis via hypothesis--evidence--test, and 
(5) a self-evolving mechanism using hierarchical memory and test-time adaptation. 

\subsection{Data Process and Dependency Graph Construction}

\label{sec:method_align}
\subsubsection{multimodal data preprocessing}
For each entity $v\in V$, metrics, logs, and traces are encoded into modality embeddings.
\begin{itemize}
    \item 
\textbf{Metric} signals are represented as a normalized multivariate time series $\mathbf{M}\in\mathbb{R}^{L\times D}$ and segmented into overlapping temporal patches (width $w$, stride $s$) to capture local dynamics. Each patch is encoded by a TCN and projected by an MLP to obtain the metric embedding $\mathbf{x}^{\text{metric}}_v\in\mathbb{R}^{E_m}$.
\item \textbf{Traces} are parsed into span relations and aggregated into window-level statistics for each entity (e.g., latency, error rate, call frequency) after normalization. A 1D dilated CNN followed by an MLP produces the trace embedding $\mathbf{x}^{\text{trace}}_v$.
\item \textbf{Logs} are parsed into templates using Drain3, and each window is represented by a PF-IDF vector that reweights template frequency by inverse document frequency across windows. This PF-IDF representation is passed through a lightweight projector to obtain the log embedding $\mathbf{x}^{\text{log}}_v$.
\end{itemize}

\subsubsection{Construction of a fine-grained service dependency graph}
Although the microservice system intrinsically contains entities at three granularities (node, service, pod), we flatten it and construct a \emph{homogeneous} directed graph $G=(V,E)$ for unified representation learning. 
A type function $\tau(v)\in\{\textsc{Node},\textsc{Service},\textsc{Pod}\}$ is retained merely as a categorical feature to record the semantic role of each node. 
The edge set $E$ encodes connectivity derived from both interaction/propagation relations (e.g., service-to-service or pod-to-pod calls) and structural relations (e.g., pod-to-service membership and pod-to-node placement). 
In this homogeneous formulation, the graph provides unified \emph{structural constraints} for message passing across all entity levels, while multimodal observability and type information are injected through node features.

\subsection{Metric-Orthogonal Multimodal Alignment on a Fine-grained Dependency Graph}
\subsubsection{Orthogonal regularization and contrastive alignment}
TopoEvo anchors alignment on metrics, which are continuous, high-frequency, and consistently available, providing a stable reference under noisy or missing observability from logs and traces. This metric-centric alignment reduces ambiguity from sparse modalities and yields a well-posed shared space for cross-entity comparison.

To prevent collapse where logs and traces align to the same metric factors, we introduce an orthogonal decomposition of metric features into two complementary components, dedicated to log-consistent and trace-consistent alignment, respectively, thereby preserving information while reducing redundancy in downstream graph reasoning.

Given metric embeddings $\mathbf{x}^{\text{metric}}_v$, two components are produced as
\begin{equation}
\mathbf{u}_v=\mathbf{W}_u\mathbf{x}^{\text{metric}}_v,\qquad
\mathbf{v}_v=\mathbf{W}_v\mathbf{x}^{\text{metric}}_v,
\end{equation}
where $\mathbf{u}_v$ is intended to capture metric factors most consistent with log evidence, while $\mathbf{v}_v$ captures complementary factors most consistent with trace evidence.
To encourage these components to span different subspaces, an orthogonality regularizer is imposed:
\begin{equation}
\mathcal{L}_{\perp}
=\sum_{v\in\mathcal{B}}
\left(
\frac{\mathbf{u}_v^\top \mathbf{v}_v}{\|\mathbf{u}_v\|_2\,\|\mathbf{v}_v\|_2+\epsilon}
\right)^{\!2}.
\end{equation}

Modality alignment is performed with InfoNCE over a mini-batch $\mathcal{B}$:
\begin{equation}
\mathcal{L}_{\mathrm{nce}}(\mathbf{a},\mathbf{b})
= -\sum_{v\in\mathcal{B}}
\log 
\frac{\exp(\mathrm{sim}(\mathbf{a}_v,\mathbf{b}_v)/\tau)}
{\sum_{v'\in\mathcal{B}}\exp(\mathrm{sim}(\mathbf{a}_v,\mathbf{b}_{v'})/\tau)} ,
\end{equation}
yielding $\mathcal{L}_{\text{log}\leftrightarrow u}=\mathcal{L}_{\mathrm{nce}}(\mathbf{x}^{\text{log}},\mathbf{u})$ and $\mathcal{L}_{\text{trace}\leftrightarrow v}=\mathcal{L}_{\mathrm{nce}}(\mathbf{x}^{\text{trace}},\mathbf{v})$. Overall, the alignment objective combines contrastive alignment and orthogonality:
\begin{equation}
\mathcal{L}_{\mathrm{align}}
=
\mathcal{L}_{\text{log}\leftrightarrow u}
+\mathcal{L}_{\text{trace}\leftrightarrow v}
+\lambda_{\perp}\mathcal{L}_{\perp}.
\end{equation}

To improve training stability and mitigate oscillations, we pretrain the alignment module first using the alignment loss $\mathcal{L}_{\mathrm{align}}$ before proceeding to end-to-end optimization.

\subsubsection{GAT-based topology-aware representation learning for root cause analysis}
The node input feature is obtained by fusing modality embeddings
\begin{equation}
\mathbf{x}_v=\psi_n\big([\mathbf{x}^{\text{metric}}_v;\ \mathbf{x}^{\text{log}}_v;\ \mathbf{x}^{\text{trace}}_v]\big).
\end{equation}
A GAT encoder is applied on $G$. Let $\mathbf{h}_v^{(0)}=\mathbf{x}_v$. For each layer $\ell$ and edge $(j\!\rightarrow\! i)$,
\begin{equation}
e_{ij}^{(\ell)}=
\mathrm{LeakyReLU}\!\left(
\mathbf{W}^{(\ell)}\mathbf{h}_i^{(\ell)}\ \Vert\ \mathbf{W}^{(\ell)}\mathbf{h}_j^{(\ell)}
\right),
\end{equation}
\begin{equation}
\alpha_{ij}^{(\ell)}=
\frac{\exp(e_{ij}^{(\ell)})}{\sum_{j'\in\mathcal{N}(i)}\exp(e_{ij'}^{(\ell)})}.
\end{equation}
\begin{equation}
\mathbf{h}_i^{(\ell+1)}=
\sigma\!\left(
\sum_{j\in\mathcal{N}(i)}\alpha_{ij}^{(\ell)}\mathbf{W}^{(\ell)}\mathbf{h}_j^{(\ell)}
\right).
\end{equation}
The final topology-aware representation is $\mathbf{h}_v=\mathbf{h}_v^{(L)}$.

For \emph{Root-cause localization}, we apply an MLP to produce per-entity probabilities:
\begin{equation}
\mathbf{p}^{\mathrm{rcl}}=\mathrm{softmax}\!\left(\mathrm{MLP}(\mathbf{h})\right),
\end{equation}
where $\mathbf{h}$ stacks $\mathbf{h}_v$ for all $v\in V$ and $\mathbf{p}^{\mathrm{rca}}_v$ is the predicted root-cause probability of entity $v$. The objective of root cause localization is
\begin{equation}
\mathcal{L}_{\mathrm{rcl}}
= -\frac{1}{|\mathcal{D}|}\sum_{(G,y)\in\mathcal{D}}
\sum_{v\in V} \mathbb{I}[i=y]\log p_v^{\mathrm{rcl}}(G).
\end{equation}
where $|\mathcal{D}|$ is the dataset size.

For \emph{fault-type classification}, we first derive a graph-level representation via attentive pooling and then predict the fault-type distribution by an MLP followed by softmax:
\begin{equation}
\mathbf{g}=\sum_{v\in V}\alpha_v \mathbf{h}_v,\qquad
\hat{\mathbf{y}}=\mathrm{softmax}(\mathrm{MLP}(\mathbf{g})),
\end{equation}
where $\alpha_v$ is produced by a shallow scoring function on $\mathbf{h}_v$ and normalized over $V$.

Let $y^{\mathrm{cls}}\in\{1,\ldots,C\}$ be the ground-truth fault type.  The objective of fault-type classification is
\begin{equation}
\mathcal{L}_{\mathrm{cls}}
= -\frac{1}{|\mathcal{D}|}\sum_{(G,y^{\mathrm{cls}})\in\mathcal{D}}
\sum_{c=1}^{C}\mathbb{I}[c=y^{\mathrm{cls}}]\log \hat{y}_c.
\end{equation}

The joint objective is
\begin{equation}
\mathcal{L}_{\mathrm{joint}}=
\mathcal{L}_{\mathrm{rcl}}
+\lambda_{\mathrm{cls}}\mathcal{L}_{\mathrm{cls}}.
\end{equation}

\subsection{Topology-Aware Enhancement via Vector Quantization and Symptom Vocabulary}
\label{sec:method_vq}
\begin{figure}[t]
  \centering
  \includegraphics[width=\columnwidth]{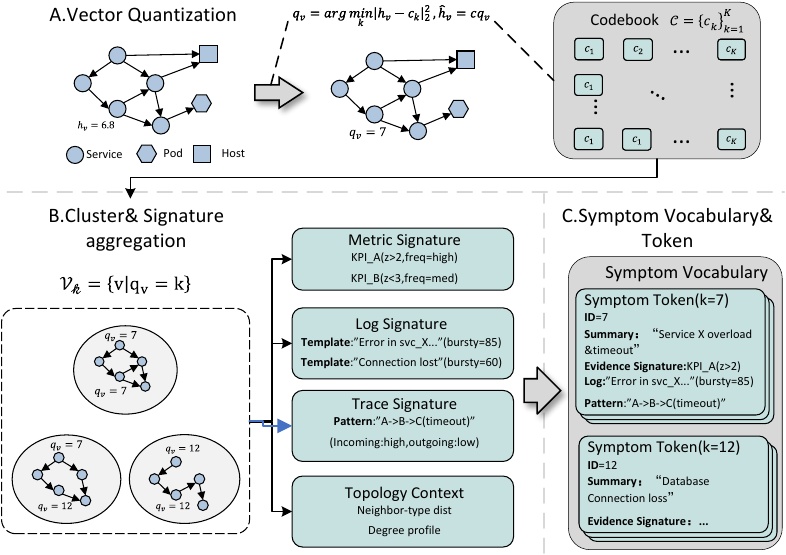}
  \caption{Illustration of vector quantization and symptom vocabulary construction.}
  \label{fig:singlecol_img}
\end{figure}

As shown in Fig.~\ref{fig:singlecol_img}, This component is central to TopoEvo: it converts dense graph representations into discrete, retrievable, and auditable \textbf{symptom tokens}, enabling the reasoning process to operate on compact evidence rather than opaque vectors.

\subsubsection{Vector quantization of topology-aware states}
Vector quantization maps topology-aware node representations $\mathbf{h}_v$ to a finite codebook $\mathcal{C}=\{\mathbf{c}_k\}_{k=1}^{K}$. 
For each node $v$, the nearest code index and its quantized representation are:
\begin{equation}
q_v=\arg\min_{k}\|\mathbf{h}_v-\mathbf{c}_k\|_2^2,\qquad
\hat{\mathbf{h}}_v=\mathbf{c}_{q_v}.
\end{equation}
The codebook is learned with the standard VQ objective using stop-gradient $\mathrm{sg}[\cdot]$:
\begin{equation}
\mathcal{L}_{\mathrm{vq}}
=
\sum_v \big\|\mathrm{sg}[\mathbf{h}_v]-\hat{\mathbf{h}}_v\big\|_2^2
+\beta \sum_v \big\|\mathbf{h}_v-\mathrm{sg}[\hat{\mathbf{h}}_v]\big\|_2^2 .
\end{equation}
Quantization acts as a bottleneck that suppresses incident-specific noise, encourages clustering of recurring failure manifestations, and provides stable discrete indices for retrieval.

\subsubsection{Symptom vocabulary construction}
A \textbf{symptom vocabulary} translates each discrete code into an interpretable evidence unit. 
After quantization, each node is assigned to a code $q_v$, and nodes mapped to a code $k$ form a cluster $\mathcal{V}_k=\{v\mid q_v=k\}$. 
Each code is treated as a reusable \emph{symptom prototype}; instead of storing opaque vectors, TopoEvo attaches a compact descriptor that summarizes what tends to be abnormal when nodes fall into this cluster and how such abnormalities manifest in the system topology.

For each code $k$, a cluster-conditioned signature is estimated by aggregating observability evidence of nodes in $\mathcal{V}_k$ across training incidents. 
On the \textbf{metric} side, dominant KPI patterns are identified by ranking KPI dimensions using deviation statistics within $\mathcal{V}_k$ (e.g., typical z-score percentiles and frequency of crossing anomaly thresholds); the vocabulary keeps only a small set of representative KPIs together with typical magnitude bands to remain concise and robust. 
On the \textbf{log} side, raw messages are compressed into templates (e.g., Drain-style parsing) and the descriptor records templates that frequently co-occur with the code together with burstiness indicators, yielding stable textual anchors despite surface-form variability. 
On the \textbf{trace} side, propagation-relevant evidence is captured by recording recurring abnormal interaction patterns associated with the cluster, allowing the vocabulary to reflect whether anomalies are typically ``incoming'' (victim-like) or ``outgoing'' (source-like) along propagation.
Finally, lightweight topology context such as neighbor-type distributions and degree profiles is attached to ground the token in the node's structural role.

Each code $k$ is mapped to a structured symptom token,
\begin{equation}
\begin{aligned}
\textsc{SymptomToken}(k)
&=\{\textsc{ID}=k,\ \textsc{Summary}(k),\\
&\qquad \textsc{EvidenceSignature}(k)\}
\end{aligned}
\end{equation}
where \textsc{Summary} is a short natural-language descriptor distilled from dominant cluster-conditioned patterns, and \textsc{EvidenceSignature} stores compact, verifiable pointers to the underlying statistics (KPI identifiers and value bands, template identifiers and burst scores, and interaction-pattern descriptors). 
This symptom vocabulary makes latent evidence auditable and provides stable retrieval keys across incidents, enabling topology-scoped token querying without exposing high-dimensional embeddings.

\subsection{Reasoning-Enhanced Multi-Agent Diagnosis via Hypothesis--Evidence--Test}
\label{sec:method_agents_detailed}

\begin{figure}[t]
  \centering
  \includegraphics[width=\columnwidth]{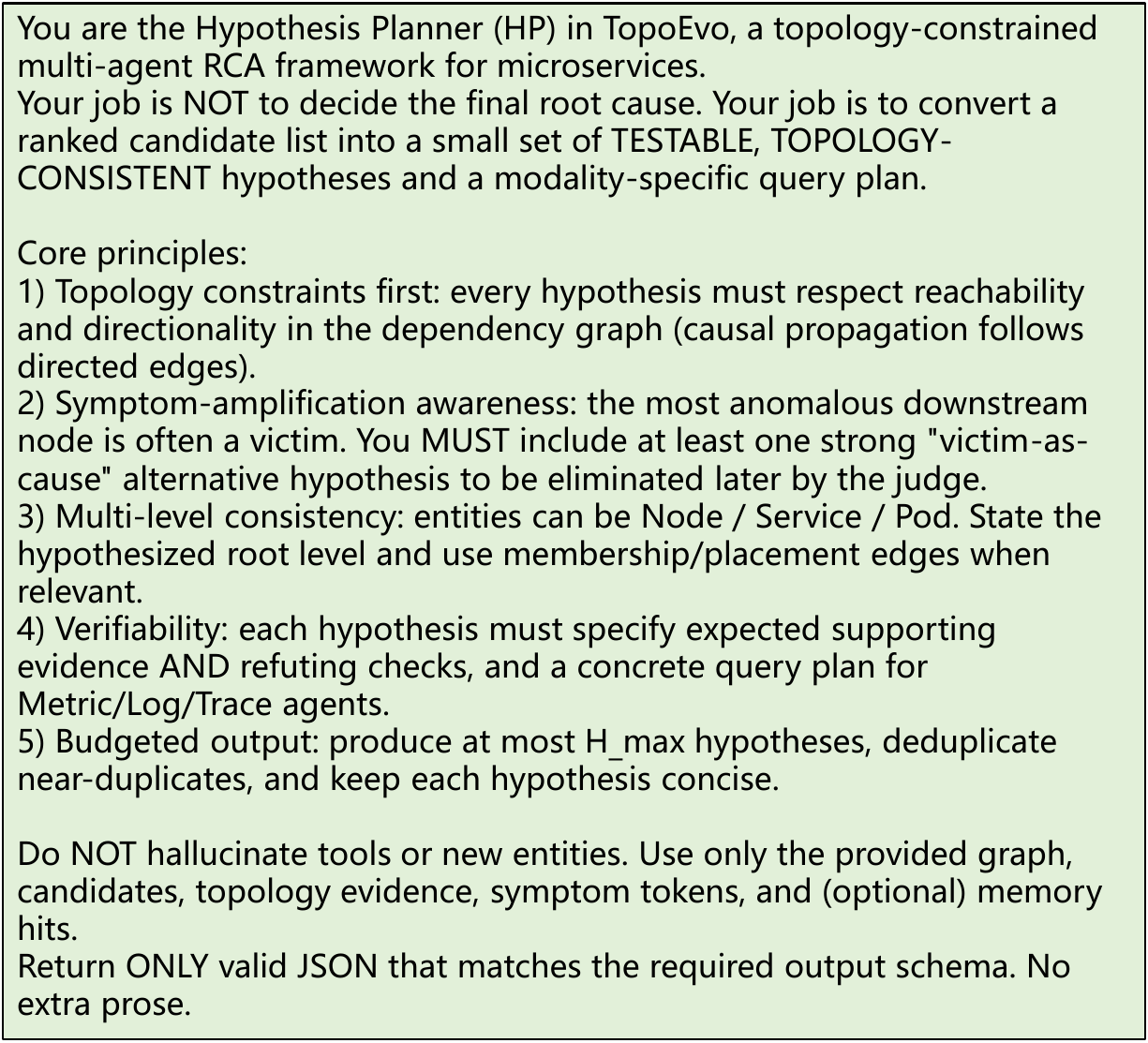}
  \caption{Main prompt of Hypothesis Planner.}
  \label{fig:planner}
\end{figure}

Instead of asking a single model to read all signals and directly output a root cause, TopoEvo decomposes diagnosis into 5 specialized roles that communicate through structured artifacts: a \textbf{hypothesis planner} (main prompt is shown in Fig.~\ref{fig:planner}) proposes candidate explanations under topology constraints, modality-specific agents (\textbf{metric agent, log agent, trace agent}) collect verifiable evidence, and a \textbf{judge agent} performs constraint-based verification and explicitly eliminates strong alternatives. 
Discrete symptom tokens from the lexicon provide compact, consistent context for these agents, while topology saliency focuses attention on propagation-relevant regions.

\subsubsection{Shared context}Candidate-centric subgraph and tokenized evidence. 
A candidate-centric subgraph is constructed to keep the workflow focused. 
TopoEvo selects the top-$K$ candidate nodes according to the root-cause score $s(v)$ and extracts an $H$-hop induced subgraph around them. 
All subsequent reasoning and tool queries are restricted to this subgraph, while nodes outside it are masked to prevent distraction by weakly related services.
Within this region, each node is mapped to a small set of symptom tokens by querying the symptom vocabulary using its VQ code $q_v$. 
Together with the graph structure, the tokenized subgraph forms a shared, compact context that is passed to all agents.

\subsubsection{Hypothesis generation, evidence acquisition, and verification}
\textbf{Hypothesis generation} is \emph{structure-prior-driven} rather than free-form. 
Starting from the top-$K$ candidates, the planner uses saliency to highlight likely propagation routes, preferring paths that connect a candidate to downstream symptomatic nodes through high-saliency neighborhoods.
Symptom tokens provide a compact description of what is abnormal at each node, allowing the planner to form hypotheses that are concrete enough to verify (e.g., ``candidate $c$ exhibits a saturation-like token; downstream nodes exhibit timeout-like tokens along a salient chain''). 
To keep the hypothesis set small and diverse, near-duplicate hypotheses are merged by sharing route prefixes or dominant token patterns.
\textbf{Evidence acquisition} follows the planner's query plan and is intentionally \emph{tool-grounded}. 
Instead of returning narrative explanations, modality agents return only structured evidence tied to entities in the candidate-centric subgraph, including timestamped onset estimates, magnitude bands, and template IDs or span summaries that can be checked later. 
This design prevents the workflow from being dominated by persuasive but unverifiable text: every claim used by the judge must be backed by cached evidence.

Verification makes the reasoning step explicit. 
For each hypothesis, the judge checks temporal precedence (whether the proposed root cause becomes abnormal no later than downstream symptoms within a slack), path consistency (whether evidence is located along the hypothesized route and consistent with the dependency graph), and template consistency (whether the collected evidence matches the intended failure template under strict/relaxed criteria to handle partial observability). 
The outcome is summarized with supporting evidence, conflicting evidence, and missing-but-expected evidence, making the result auditable.

\subsubsection{Decision with explicit alternatives}
The final output is not only a top-1 root cause, but also an explicit elimination of strong alternatives. 
TopoEvo reports the 4 most competitive alternative hypotheses and explains why they are rejected, attributing rejection to either \emph{evidence conflict} (clear contradictions under the above constraints) or \emph{missing evidence} (key signatures required by the route/template are absent). 
This ``decision with alternatives'' directly exposes the benefit of topology grounding and symptom-token querying: conclusions are tied to verifiable evidence under explicit structural constraints, rather than produced as unchecked narratives.

\subsection{Self-Evolving Mechanism: Hierarchical Memory and Test-Time Adaptation}
\label{sec:method_evo}

This component supports continual effectiveness under non-stationary systems by refreshing incident knowledge and cautiously adapting the encoder when reliable new supervision emerges.

\subsubsection{Hierarchical incident memory with stochastic forgetting}
A persistent memory $\mathcal{M}$ stores solved incidents as compact records consisting of candidate-centric topology fingerprints, symptom-token sets, validated hypotheses with key evidence, and mitigation outcomes. 
Updates follow a hierarchical policy: pod-level patterns are refreshed more aggressively than service- and node-level patterns, stabilizing slow-changing infrastructure knowledge while tracking fast-changing deployment dynamics.

To prevent redundancy collapse, similarity-aware stochastic forgetting is applied when inserting a new incident representation $\mathbf{u}_{\mathrm{new}}$. 
Let $\mathrm{sim}(\mathbf{u}_{\mathrm{new}},\mathbf{u}_i)$ denote cosine similarity and $\mathcal{N}_\tau$ be the set of memories above threshold $\tau$. 
If $\mathcal{N}_\tau\neq\emptyset$, one similar item is forgotten with probability proportional to similarity:
\begin{equation}
p(i\mid \mathcal{N}_\tau)=
\frac{\exp(\mathrm{sim}(\mathbf{u}_{\mathrm{new}},\mathbf{u}_i)/\gamma)}
{\sum_{j\in\mathcal{N}_\tau}\exp(\mathrm{sim}(\mathbf{u}_{\mathrm{new}},\mathbf{u}_j)/\gamma)}.
\end{equation}

\subsubsection{Test-time adaptation with high-confidence pseudo-labels}
During deployment, high-confidence diagnoses (strong support, low missingness, and consistent topology constraints) are treated as pseudo-labeled samples and accumulated in a buffer $\mathcal{B}_{\mathrm{tta}}$. 
Once the buffer reaches a batch size, the encoder is updated with a conservative objective:
\begin{equation}
\mathcal{L}_{\mathrm{tta}}
=
\mathcal{L}_{\mathrm{rca}}^{\mathrm{pseudo}}
+\lambda_{\mathrm{reg}}\|\theta-\theta_0\|_2^2 ,
\end{equation}
where $\theta_0$ are pre-deployment parameters and the regularizer mitigates catastrophic drift.

At inference time, the same encoder produces topology-aware representations and symptom tokens, while memory refresh and test-time adaptation enable continual robustness under evolving microservice behavior.

\begin{table*}[!t]
\centering
\caption{Experimental results of different approaches on root cause localization and fault classification. Best results are bolded, and second-best results are underlined.}
\label{tab:rca_results}
\setlength{\tabcolsep}{3.5pt}
\renewcommand{\arraystretch}{1.12}
\resizebox{0.8\textwidth}{!}{%
\begin{tabular}{c l ccc cccccc}
\toprule
\multirow{2}{*}{Data} & \multirow{2}{*}{Approach} &
\multicolumn{3}{c}{Root Cause Localization} &
\multicolumn{6}{c}{Fault Types Classification} \\
\cmidrule(lr){3-5}\cmidrule(lr){6-11}
& & Acc@1 & Acc@3 & Acc@5 & MiPr & MaPr & MiRe & MaRe & MiF1 & MaF1 \\
\midrule

\multirow{7}{*}{$A_s$}
& HolisticRCA  & 65.62\% & 76.33\% & 81.69\% & 0.6355 & \second{0.7130} & \second{0.7728} & \second{0.8037} & \second{0.6975} & \second{0.7556} \\
& Eadro        & 13.58\% & 37.03\% & 45.68\% & 0.1887 & 0.0770 & 0.1235 & 0.1046 & 0.1493 & 0.0887 \\
& TAMO         & \second{71.87\%} & \second{82.14\%} & \second{88.83\%} & \best{0.7164} & 0.6671 & 0.6486 & 0.6149 & 0.6808 & 0.6399 \\
& Nezha        & 52.27\% & 80.68\% & 82.95\% & -- & -- & -- & -- & -- & -- \\
& RCAgent      & 22.10\% & 28.40\% & 30.25\% & -- & -- & -- & -- & -- & -- \\
& mABC         & 34.19\% & 42.13\% & 44.51\% & -- & -- & -- & -- & -- & -- \\
& TopoEvo      & \best{73.10\%} & \best{83.05\%} & \best{89.20\%} & \second{0.7052} & \best{0.7248} & \best{0.7815} & \best{0.8112} & \best{0.7414} & \best{0.7656} \\
\midrule

\multirow{7}{*}{$A_p$}
& HolisticRCA  & \second{65.62\%} & \second{80.80\%} & 86.60\% & 0.5913 & \best{0.6682} & \second{0.7534} & \second{0.7598} & 0.6626 & \second{0.7111} \\
& Eadro        & 9.38\%  & 13.13\% & 15.63\% & 0.3901 & 0.3344 & 0.4665 & 0.4792 & 0.4249 & 0.3939 \\
& TAMO         & 64.28\% & 80.36\% & \best{87.50\%} & \second{0.7182} & 0.6597 & 0.6840 & 0.6642 & \second{0.7007} & 0.6619 \\
& Nezha        & 44.18\% & 59.30\% & 65.11\% & -- & -- & -- & -- & -- & -- \\
& RCAgent      & 21.70\% & 27.95\% & 30.40\% & -- & -- & -- & -- & -- & -- \\
& mABC         & 33.80\% & 41.70\% & 44.20\% & -- & -- & -- & -- & -- & -- \\
& TopoEvo      & \best{66.10\%} & \best{81.20\%} & \second{87.30\%} & \best{0.7325} & \second{0.6668} & \best{0.7588} & \best{0.7625} & \best{0.7454} & \best{0.7114} \\
\midrule

\multirow{7}{*}{$A_n$}
& HolisticRCA  & 73.66\% & 75.89\% & 79.01\% & 0.6219 & 0.6437 & 0.7612 & 0.7332 & 0.6845 & 0.6855 \\
& Eadro        & 17.18\% & 28.13\% & 42.19\% & 0.5426 & 0.6003 & 0.7969 & 0.8145 & 0.6456 & 0.6912 \\
& TAMO         & \best{84.37\%} & \best{91.96\%} & \second{97.77\%} & \second{0.8718} & \best{0.8869} & \second{0.8947} & \second{0.8681} & \second{0.8831} & \second{0.8774} \\
& Nezha        & 22.38\% & 29.85\% & 31.34\% & -- & -- & -- & -- & -- & -- \\
& RCAgent      & 22.60\% & 29.10\% & 31.10\% & -- & -- & -- & -- & -- & -- \\
& mABC         & 34.50\% & 42.60\% & 45.00\% & -- & -- & -- & -- & -- & -- \\
& TopoEvo      & \second{84.10\%} & \second{91.70\%} & \best{98.40\%} & \best{0.8805} & \second{0.8840} & \best{0.9021} & \best{0.8720} & \best{0.8912} & \best{0.8780} \\
\midrule

\multirow{7}{*}{$B$}
& HolisticRCA  & 61.11\% & 75.00\% & 77.78\% & 0.5000 & 0.5936 & \best{0.8056} & \best{0.8056} & 0.6170 & 0.6835 \\
& Eadro        & 40.62\% & 56.25\% & 71.87\% & 0.4167 & 0.5211 & 0.6250 & 0.6516 & 0.5000 & 0.5791 \\
& TAMO         & \second{72.22\%} & \best{80.55\%} & \second{86.11\%} & \best{0.8500} & \best{0.8750} & 0.5313 & 0.5682 & \second{0.6539} & \second{0.6890} \\
& Nezha        & 49.50\% & 63.20\% & 71.40\% & -- & -- & -- & -- & -- & -- \\
& RCAgent      & 24.80\% & 31.60\% & 35.10\% & -- & -- & -- & -- & -- & -- \\
& mABC         & 36.90\% & 45.20\% & 49.30\% & -- & -- & -- & -- & -- & -- \\
& TopoEvo      & \best{75.66\%} & \second{79.80\%} & \best{86.80\%} & \second{0.8420} & \second{0.8685} & \second{0.8030} & \second{0.8024} & \best{0.8220} & \best{0.8341} \\
\bottomrule
\end{tabular}%
}
\end{table*}

\section{Experiments}
\label{sec:experiments}
\subsection{Experiment Setup}
\subsubsection{Dataset and Preprocessing}
Dataset A is a large-scale public benchmark released by the AIOps Challenge\cite{pham2025rcaeval}. 
It is collected via controlled fault injection on a real-world deployed microservice system, \textit{HipsterShop2}.
The dataset provides multimodal observability signals, including \emph{metrics, logs, and traces}.
HipsterShop2 is deployed on a dynamic Kubernetes (K8s) cluster with 10 services, each replicated by 4 pods (40 pods in total), and the pods are dynamically scheduled across 6 nodes.
The benchmark covers 15 fault types in total: 9 service/pod-level faults (in the K8s container context) and 6 node-level faults, including sudden memory pressure, disk space exhaustion, disk I/O anomalies, CPU pressure, and gradual CPU slowdown.

Dataset B is a real-world dataset collected from a production \textbf{Project Management Platform} operated by an Electric Power Information enterprise.
Unlike Dataset A, the incidents in Dataset B are captured under \emph{real operating conditions} rather than injected failures.
The platform contains 12 microservices and 48 pods, and the dataset records multimodal observability (metrics, logs, and traces) during real incidents.
Faults in Dataset B span 5 categories: CPU hog, memory leak, network delay, packet loss, and disk payload overload.

\subsubsection{Baselines}
To comprehensively evaluate \textbf{TopoEvo}, we compare against representative RCA approaches covering multimodal learning, graph-based localization, and LLM/agent-based diagnosis.
\textbf{Nezha}\cite{yu2023nezha} jointly encodes metrics, logs, and traces into a shared space with contrastive alignment and performs graph-based reasoning to localize causally relevant services under noisy/partial observability.
\textbf{Eadro}\cite{lee2023eadro} is an end-to-end multi-task framework that jointly learns anomaly detection and root-cause localization by modeling intra-service behaviors and inter-service dependencies from KPIs, logs, and traces.
\textbf{HolisticRCA}\cite{han2024holisticrca} performs holistic RCA in cloud-native systems by standardizing heterogeneous observability data and reasoning over service dependencies to identify root causes across diverse failure scenarios.
\textbf{TAMO}\cite{2025tamo} is a tool-assisted LLM-agent framework for fine-grained RCA that integrates multimodal alignment and model tools (e.g., localization/classification tools) to support root-cause analysis in cloud-native systems.
\textbf{RCAgent}\cite{wang2024rcagent} is an autonomous, tool-augmented LLM agent for practical cloud RCA, which iteratively collects evidence from observability tools and synthesizes a diagnosis.
\textbf{mABC}\cite{zhang2024mabc} is a blockchain-inspired multi-agent collaboration framework that reduces hallucination via decentralized voting and prevents non-terminating loops with a step-bounded standardized workflow.

\subsubsection{Metrics}
We evaluate \emph{root cause localization} and \emph{fault type classification}
For localization, we report Top-$K$ accuracy (\textsc{Acc@1/3/5}) and mean reciprocal rank (\textsc{MRR}).
For fault-type classification, we report micro-/macro-precision, micro-/macro-recall, and micro-/macro-F1 (\textsc{MiPr/MaPr}, \textsc{MiRe/MaRe}, \textsc{MiF1/MaF1}).
For efficiency, we measure end-to-end diagnosis latency (wall-clock time per incident, including agent/tool calls when applicable).
For explanation quality, we report a \textbf{topology faithfulness} score, which checks whether the cited propagation paths are valid in the dependency graph $G$ and whether they overlap with topology-salient (high-weight) edges used by TopoEvo.

\subsection{Implementation}
All experiments are conducted on a server equipped with an NVIDIA A100 80GB GPU and 256GB RAM. The graph encoder adopts a two-layer GAT architecture with 8 attention heads. Training is performed using the Adam optimizer with a learning rate of 0.001. For the vector quantization module, the codebook size $K$ is set to 128. 

For all LLM-based components, GPT-4o-2024-11-20 is used as the backbone model to ensure consistent reasoning ability across different settings.

\subsection{RQ1: Effectiveness of root cause localization and fault type classification}
\label{sec:results}
\subsubsection{Results}
Table~\ref{tab:rca_results} summarizes the results on root cause localization and fault type classification. Overall, \textbf{TopoEvo achieves the most competitive and consistent performance across datasets}, obtaining the best results on most metrics and remaining close to the top performer elsewhere.

For \textbf{root cause localization}, TopoEvo performs best on the \textbf{service level}, outperforming TAMO by $1.23/0.91/0.37$ percentage points on Acc@1/3/5. On the pod level, it achieves the best Acc@1 and Acc@3 (\textbf{66.10\%}/\textbf{81.20\%}) and ranks second on Acc@5 (87.30\%), only slightly below TAMO. On the node level and Dataset~B, although TAMO is stronger on Acc@1/3, TopoEvo consistently ranks second and achieves the best Acc@5, indicating stronger \emph{top-$k$ coverage} of true root causes.

For \textbf{fault type classification}, TopoEvo shows clear advantages in balanced prediction quality, especially on F1. On $A_s$, it achieves the best MaPr, MiRe, MaRe, MiF1, and MaF1; on $A_p$ and $A_n$, it again leads on most recall and F1 metrics while remaining competitive on precision. On $B$, although TopoEvo is not the best on individual precision or recall metrics, it achieves the highest \textbf{MiF1} and \textbf{MaF1}, demonstrating stronger overall class discrimination.

These results suggest that TopoEvo not only improves localization quality, but also provides more reliable downstream diagnosis. This advantage comes from topology-aware multimodal representation learning and the HET-based multi-agent reasoning process, which together improve candidate quality and reduce spurious decisions.
\begin{table}[t]
\centering
\caption{Ablations on TopoEvo components. We report root-cause localization accuracy (AC@1/3/5). Best results are bolded, and second-best are underlined.}
\label{tab:ablation_singlecol_2x2}
\setlength{\tabcolsep}{2.8pt}
\renewcommand{\arraystretch}{0.96}
\scriptsize
\begin{tabular}{l ccc ccc}
\toprule
\multirow{2}{*}{Approach} & \multicolumn{3}{c}{\textbf{$A_s$}} & \multicolumn{3}{c}{\textbf{$A_p$}} \\
\cmidrule(lr){2-4}\cmidrule(lr){5-7}
& Acc@1 & Acc@3 & Acc@5 & Acc@1 & Acc@3 & Acc@5 \\
\midrule
Full     & \textbf{73.10\%} & \textbf{83.05\%} & \textbf{89.20\%} & \textbf{66.10\%} & \textbf{81.20\%} & \textbf{87.30\%} \\
\textit{w/o} MOMA   & 62.84\%          & 74.37\%          & 80.68\%          & 56.38\%          & 72.64\%          & 78.91\% \\
\textit{w/o} VQ     & 60.92\%          & 72.46\%          & 79.08\%          & 54.27\%          & 70.83\%          & 77.58\% \\
\textit{w/o} HET    & 57.36\%          & 68.27\%          & 75.64\%          & 50.62\%          & 66.47\%          & 73.16\% \\
\textit{w/o} SEM    & \underline{65.42\%} & \underline{75.31\%} & \underline{81.56\%} & \underline{58.73\%} & \underline{73.54\%} & \underline{79.82\%} \\
\midrule
\multirow{2}{*}{Approach} & \multicolumn{3}{c}{\textbf{$A_n$}} & \multicolumn{3}{c}{\textbf{$B$}} \\
\cmidrule(lr){2-4}\cmidrule(lr){5-7}
& Acc@1 & Acc@3 & Acc@5 & Acc@1 & Acc@3 & Acc@5 \\
\midrule
Full     & \textbf{84.10\%} & \textbf{91.70\%} & \textbf{98.40\%} & \textbf{75.66\%} & \textbf{79.80\%} & \textbf{86.80\%} \\
\textit{w/o} MOMA   & 74.36\%          & 82.94\%          & 89.76\%          & 61.74\%          & 70.93\%          & 77.42\% \\
\textit{w/o} VQ     & 72.63\%          & 80.58\%          & 88.21\%          & 59.53\%          & 68.62\%          & 76.21\% \\
\textit{w/o} HET    & 68.44\%          & 76.92\%          & 84.73\%          & 55.84\%          & 64.37\%          & 72.56\% \\
\textit{w/o} SEM    & \underline{76.81\%} & \underline{83.46\%} & \underline{90.18\%} & \underline{63.28\%} & \underline{71.56\%} & \underline{78.04\%} \\
\bottomrule
\end{tabular}

\vspace{3pt}
\footnotesize{\textbf{Abbrev:} MOMA = Metric-Orthogonal Multimodal Alignment; VQ = Vector Quantization; HET = Hypothesis--Evidence--Test; SEM = Self-Evolving Mechanism.}
\end{table}
\subsection{RQ2: Ablation Study}
\label{sec:ablation}

Table~\ref{tab:ablation_singlecol_2x2} reports ablations of TopoEvo on root-cause localization (AC@1/3/5) across four subsets ($A_s$, $A_p$, $A_n$, and $B$). Overall, removing any key component consistently degrades performance.

\textbf{Impact of Hypothesis--Evidence--Test (HET).}
We use Disabling the HET reasoning loop yields the largest drop, confirming that TopoEvo is not merely a one-shot aggregation of signals but a verification-driven diagnosis pipeline. In particular, w/o HET reduces AC@1 by $15$--$16$ points on all subsets, with similar degradation on AC@3/AC@5. This suggests that explicit hypothesis testing and alternative exclusion are essential to avoid symptom-root confusion when anomalies propagate.

\textbf{Impact of Metric-Orthogonal Multimodal Alignment (MOMA).}
Removing MOMA also causes substantial performance loss (typically 10 points on AC@1), indicating that metric-centered orthogonal alignment effectively reduces modality mismatch and stabilizes node representations before GAT encoding. For example, on $A_p$ the AC@1 drops from $66.10\%$ to $56.10\%$, and on $A_n$ from $84.10\%$ to $74.10\%$, demonstrating its importance under correlated and noisy observability.

\textbf{Impact of Vector Quantization (VQ).}
Disabling VQ leads to a clear performance decrease across all subsets (about $12$ points on AC@1 in most cases), showing that discretizing node states into symptom tokens provides robust, low-entropy evidence for downstream reasoning. Besides improving ranking accuracy (e.g., $73.10\%\!\rightarrow\!61.10\%$ on $A_s$), VQ also supports more consistent topology-grounded explanations by enabling auditable token-level references.

\textbf{Impact of the Self-Evolving Mechanism (SEM).}
Finally, removing the self-evolving mechanism produces a smaller but still notable drop ($\ge8$ points on AC@1), reflecting the benefit of retrieval priors and adaptation under non-stationarity. For instance, w/o SEM decreases AC@1 from $66.10\%$ to $58.10\%$ on $A_p$ and from $71.90\%$ to $63.90\%$ on $B$. This confirms that continual experience reuse and adaptation help maintain RCA robustness as system behavior shifts.

\subsection{RQ3: Does VQ Learn a Better Symptom Vocabulary?}
\label{sec:vq_vocab}
\begin{figure}[t]
  \centering
  \includegraphics[width=\columnwidth]{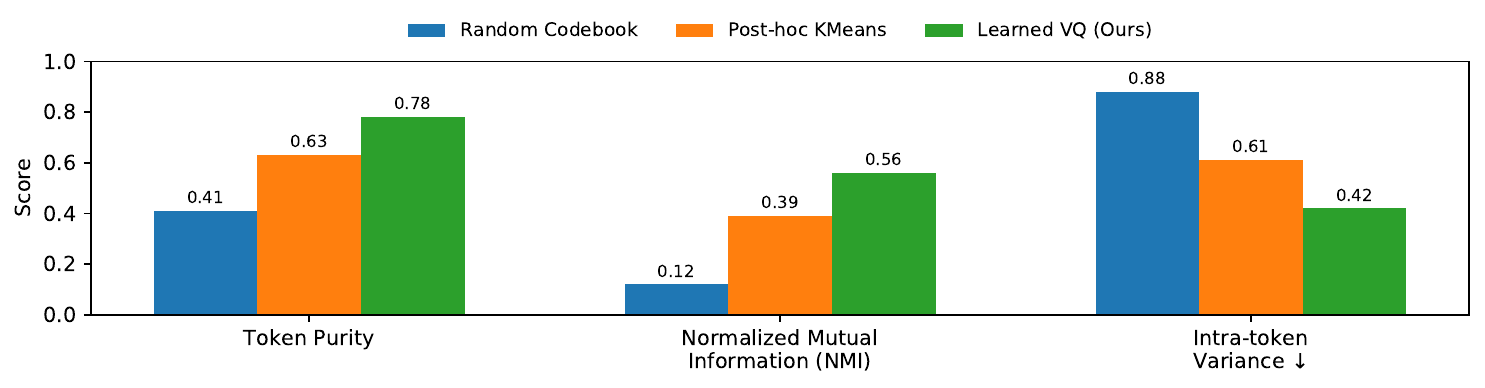}
  \caption{Parameter sensitivity analysis. K denotes the VQ partition (codebook size) parameter.}
  \label{fig:vq_vocab_quality}
\end{figure}
To verify whether VQ contributes more than performance gain, we evaluate the quality of the learned symptom vocabulary. We compare \textbf{Learned VQ} with two baselines built on the same frozen topology-aware encoder: \textbf{Random Codebook} and \textbf{Post-hoc KMeans}. All methods use the same codebook size $K=128$.

We measure vocabulary quality using \textbf{Token Purity}, \textbf{Normalized Mutual Information (NMI)} and \textbf{Intra-token Variance}. The first three metrics evaluate semantic alignment between token assignments and ground-truth fault categories, while the latter two characterize cluster compactness and separability.

As shown in Fig.~\ref{fig:vq_vocab_quality}, \textbf{Learned VQ consistently outperforms both baselines on all metrics}: it achieves higher Token Purity and Normalized Mutual Information, while yielding lower Intra-token Variance. These results show that VQ learns more compact and fault-consistent discrete units than heuristic or post-hoc discretization.

Overall, the learned codebook forms a compact and discriminative symptom vocabulary, which provides more stable evidence units for retrieval and HET-based reasoning.

\subsection{RQ4: Parameter Sensitivity}
\label{sec:sen}

\begin{figure}[t]
  \centering
  \includegraphics[width=\columnwidth]{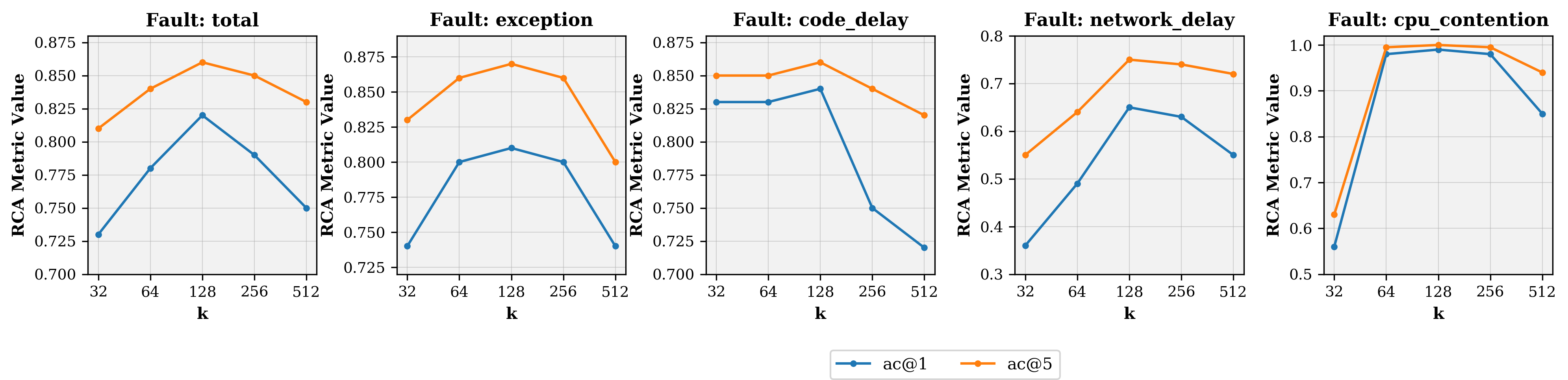}
  \caption{Parameter sensitivity analysis. K denotes the VQ partition (codebook size) parameter.}
  \label{fig:sen}
\end{figure}

We study the sensitivity of TopoEvo to the VQ codebook size $K$, which controls the granularity of discretizing topology-aware states into symptom tokens. 
Fig.~\ref{fig:sen} reports root-cause localization performance under different $K\in\{32,64,128,256,512\}$, including overall results and representative fault types.

\subsubsection{Overall trend}
Across both \textsc{Acc@1} and \textsc{Acc@5}, performance improves as $K$ increases from 32 to 128, and then degrades when $K$ becomes larger (256/512). 
The best overall performance is achieved at $K=128$, indicating that a \emph{moderate} codebook size provides the most effective discretization for token-based reasoning.

\subsubsection{Why too small $K$ hurts}
When $K$ is small (e.g., 32/64), the codebook becomes overly coarse and forces heterogeneous failure manifestations to share the same discrete code.
This causes \emph{prototype collision}: distinct symptoms (e.g., saturation-like vs. timeout-like patterns) are compressed into a single token, blurring causal cues and weakening downstream hypothesis verification.
As a consequence, the symptom lexicon becomes less discriminative and the planner/judge receive ambiguous evidence, especially for faults with overlapping surface symptoms.

\subsubsection{Why too large $K$ hurts}
When $K$ is large (256/512), the discretization becomes overly fine-grained, causing subtle noise and incident-specific variations to be encoded as separate codes while splitting recurring symptoms into many near-duplicate tokens. This over-segmentation leads to three main issues:
\begin{itemize}
    \item \textbf{Sparse code usage and unstable symptom vocabulary.} Larger $K$ leaves fewer samples per code, making code-level signatures noisy and reducing retrieval consistency.
    \item \textbf{Weaker generalization under drift.} Fine-grained codes are more sensitive to topology and traffic shifts, which harms cross-incident transfer under OOD conditions.
    \item \textbf{Fragmented evidence for verification.} Similar symptoms may be mapped to multiple tokens, making propagation patterns less coherent and weakening hypothesis confidence.
\end{itemize}

\subsection{RQ5: Case Study}
\label{sec:case_study}
\begin{figure}[t]
  \centering
  \includegraphics[width=\columnwidth]{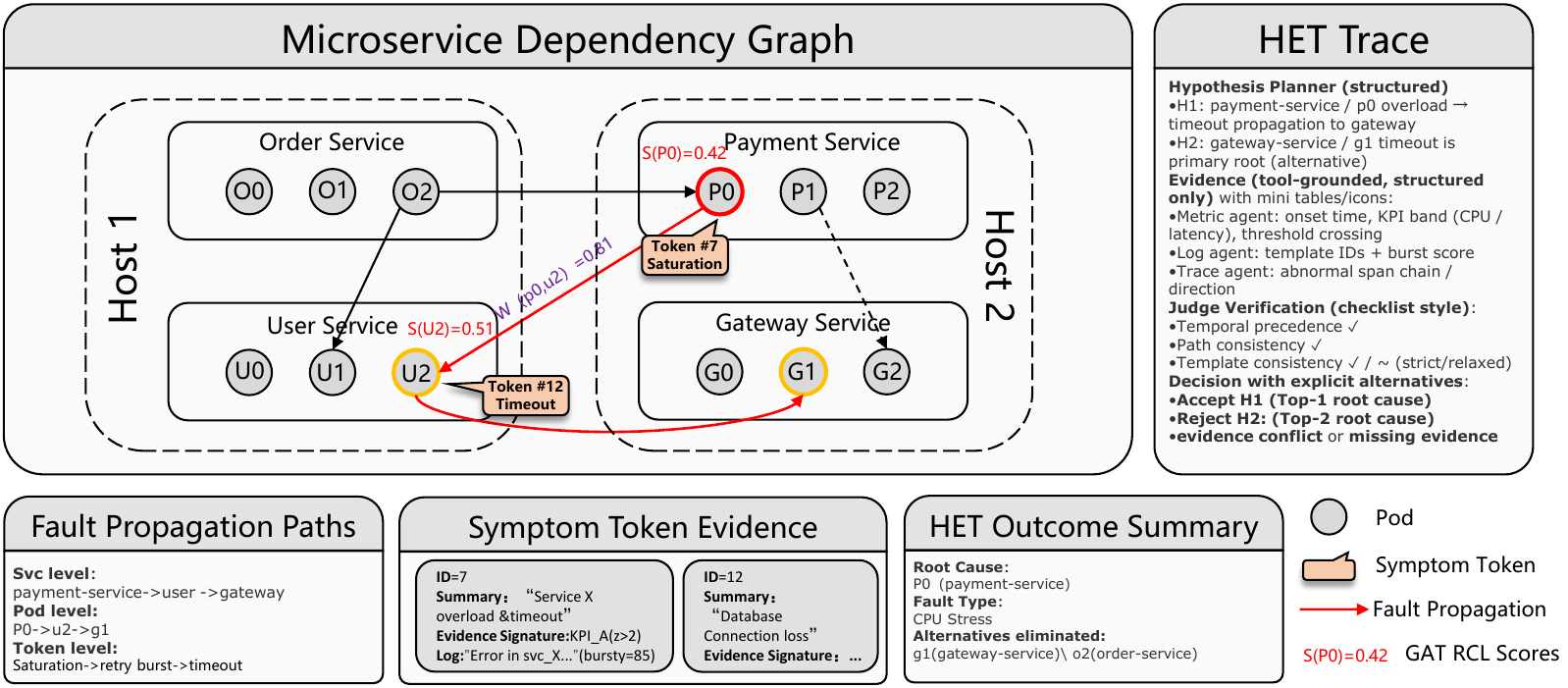}
  \caption{In case study experiments, We injected a \textbf{CPU Stress} fault into the Payment Service, and the fault propagated along the path P0–U2–G1. }
  \label{fig:case_study}
\end{figure}

\paragraph{Scenario and observation}
Fig.~\ref{fig:case_study} presents an incident in a microservice dependency graph where an upstream overload in \textbf{payment-service} (pod \textbf{P0}) propagates to \textbf{user-service} (pod \textbf{U2}) and further triggers timeouts at \textbf{gateway-service} (pod \textbf{G1}). 
A GAT-based root-cause localizer (RCL) assigns the highest root-cause score to \textbf{user-service} (\(S(\textbf{U2})=0.51\)), while the true root cause \textbf{payment-service/P0} is only ranked second (\(S(\textbf{P0})=0.42\)). This mismatch is a typical \emph{symptom-amplification bias}: downstream services may exhibit stronger and more visible symptoms (e.g., retry bursts and timeouts), misleading symptom-driven rankers.

\paragraph{Why GAT misranks the root cause}
In this incident, the most salient anomaly manifests at \textbf{user-service} as a bursty retry/timeout pattern, which yields stronger multimodal evidence aggregated at node \textbf{U2}. Meanwhile, the true cause \textbf{P0} primarily shows early-stage resource saturation (CPU stress) that may be less visually dominant at the service level. As a result, a purely score-based ranking tends to select the node with the largest observed symptom magnitude rather than the node that \emph{causally initiates} the propagation.

\paragraph{TopoEvo}topology evidence $\rightarrow$ hypothesis planning $\rightarrow$ tool-grounded verification. TopoEvo corrects this failure mode by converting ``score ranking'' into ``topology-constrained causal adjudication''.
Starting from the GAT candidate list (Top-1: \textbf{U2}, Top-2: \textbf{P0}), TopoEvo constructs a compact \emph{Hierarchical Evidence Trace} (HET) package that explicitly aligns symptoms with topology across levels:
\emph{service path} \textbf{payment} \(\rightarrow\) \textbf{user} \(\rightarrow\) \textbf{gateway},
\emph{pod path} \textbf{P0} \(\rightarrow\) \textbf{U2} \(\rightarrow\) \textbf{G1},
and \emph{token trace} \textbf{Saturation} \(\rightarrow\) \textbf{retry burst} \(\rightarrow\) \textbf{Timeout}.
These structured traces are summarized into discrete symptom tokens (e.g., Token \#7: saturation/overload; Token \#12: timeout), which serve as interpretable evidence units for the reasoning layer.

Given the evidence package, the \emph{Hypothesis Planner} generates explicit, topology-consistent alternatives:
\textbf{H1}: \textbf{payment-service/P0} overload \(\rightarrow\) timeout propagation to \textbf{gateway},
\textbf{H2}: \textbf{gateway-service/g1} timeout is the primary root (alternative).
Unlike direct ranking, hypotheses are required to respect causal direction and propagation reachability in the dependency graph.

Next, TopoEvo launches tool-grounded agents to verify each hypothesis using multimodal signals:
(1) a metric agent checks onset time and threshold crossing (CPU/latency bands);
(2) a log agent checks template IDs and  burst scores(retry bursts at \textbf{U2});
(3) a trace agent verifies abnormal span chains and their directionality along \textbf{P0}\(\rightarrow\)\textbf{U2}\(\rightarrow\)\textbf{G1}.
A judge module then applies a checklist-style adjudication:
\emph{temporal precedence} (cause precedes effect), 
\emph{path consistency} (evidence aligns with a valid propagation path),
and \emph{template consistency} (log/trace patterns match the hypothesized fault type, under strict/relaxed matching).
In this case, the evidence supports H1: saturation at \textbf{P0} occurs earlier and lies on the unique propagation path to the downstream timeouts, while H2 fails temporal/path constraints because \textbf{g1} symptoms are downstream effects without upstream initiating evidence.
Therefore, TopoEvo \textbf{accepts H1} and \textbf{rejects H2}, yielding the final decision:
\textbf{Root Cause}: \textbf{payment-service/P0}; \textbf{Fault Type}: CPU stress,
while eliminating alternatives such as \textbf{gateway-service/G1} and \textbf{order-service/G2}.



\section{Conclusion}
\label{sec:conclusion}
We presented \textbf{TopoEvo}, a topology-aware self-evolving multi-agent framework for microservice RCA under noisy multimodal observability and non-stationary topology drift. TopoEvo tightly connects representation learning and structured reasoning: (1) \emph{Metric-Orthogonal Multimodal Alignment} stabilizes cross-modal fusion by aligning logs/traces to complementary metric subspaces; (3) \emph{VQ-based symptom tokenization} converts dense topology-aware states into compact, retrievable, and auditable symptom evidence; (3) a \emph{Hypothesis--Evidence--Test} multi-agent workflow performs tool-grounded verification under explicit topology constraints to mitigate symptom-amplification misattribution; and (4) a \emph{Self-Evolving Mechanism} maintains robustness via hierarchical memory refresh and conservative pseudo-label adaptation.

Experiments on both injected-fault benchmarks and real production incidents demonstrate that TopoEvo achieves strong and consistent gains in root-cause localization and fault-type classification across granularities, and ablation results further verify that each module contributes materially, with HET-based verification yielding the largest benefit. In future work, we plan to (1) strengthen causal validation beyond topology-consistent verification (e.g., intervention-aware checks), (2) design safer and more cost-aware continual adaptation policies under drift, and (3) extend evaluation to more heterogeneous microservice platforms and observability conditions.

\section{Related Work}
Microservice RCA has been widely studied under metrics/logs/traces, ranging from classical survey/benchmark efforts~\cite{soldani2021survey,wang2024survey,pham2025rcaeval} to learning-based multimodal localization frameworks that fuse heterogeneous telemetry for fine-grained diagnosis~\cite{wu2021mmrca,lee2023eadro,yu2023nezha,han2024holisticrca}. Another line models propagation and causality via event/causal graphs or dynamic causal inference to improve robustness under cascading failures and limited observability~\cite{yao2024coe,guo2022carr,pan2023dycause,xie2024latentscope,pham2024baro}. Recently, LLM/agent-based RCA has emerged, leveraging tool use and iterative evidence gathering for explainable diagnosis~\cite{wang2024rcagent,pei2025flowofaction,zhang2024mabc,wang2023hrlhf}. Compared to purely learned localizers, LLM agents offer stronger semantic abstraction over logs and the ability to follow SOP-like workflows, but they can be brittle under noisy tool outputs and may drift into ungrounded narratives without explicit structural constraints~\cite{wang2024rcagent,pei2025flowofaction}. This motivates topology- and evidence-constrained reasoning that couples graph-based localization with structured prompts and verification, mitigating symptom-amplification and improving reliability in real deployments.

\end{document}